\documentclass[11pt,a4paper]{article}
\usepackage[hyperref]{acl2019}
\usepackage{times}
\usepackage{latexsym}
\usepackage{amssymb}
\usepackage{booktabs}
\usepackage{graphicx}

\usepackage{url}

\usepackage{mathrsfs}

\aclfinalcopy 

\title{The Effect of Downstream Classification Tasks\\ for Evaluating Sentence Embeddings}

\author{Peter Potash \\
  Microsoft Research Montreal \\
  \texttt{peter.potash@microsoft.com}\\}

\date{}

\begin{document}
\maketitle
\begin{abstract}
One popular method for quantitatively evaluating the utility
of sentence embeddings involves using them in downstream
language processing tasks that require sentence representations
as input. One simple such task is classification, where the
sentence representations are used to train and test models
on several classification datasets. We argue that by
evaluating sentence representations in such a manner,
the goal of the representations becomes learning a
low-dimensional factorization of a sentence-task label
matrix. We show how
characteristics of this matrix can affect the ability
for a low-dimensional factorization to perform as sentence
representations in a suite of classification tasks.
Primarily, sentences that have more labels across
all possible classification tasks have a higher
reconstruction loss, however the general nature of this effect 
is ultimately dependent on the overall distribution of labels
across all possible sentences.
\end{abstract}

\section{Introduction}
\label{sec:intro}

Universal Sentence Embeddings (USEs) is a line of research that seeks
to learn a mapping between token sequences and vectors
in a real-valued space of a fixed dimension.
The work of \newcite{conneau2017supervised} has brought
tremendous attention to the problem of USEs,
not only because of the model they propose, but because of
\textit{how} they propose to test the utility of their
learned representations. The authors propose to use their
learned sentence embeddings to train and test classification
models on a variety of datasets, while keeping the sentence
representation model fixed \textendash \, the suite of
downstream tasks is called SentEval \cite{conneau2018senteval} by the authors.

The research community has come to embrace SentEval as the
default method for evaluating sentence embeddings, unleashing
numerous papers that evaluate with SentEval (c.f.
\citet{subramanian2018learning,pagliardini2018unsupervised,logeswaran2018efficient,zhang2018learning}).
In our work, we seek to analyze the implications
that arise from evaluating USEs using downstream
classification tasks, namely by viewing USEs as
a low-rank approximation of a sentence-task label
matrix. Furthermore, we provide in an initial exploration
of how the distribution of task labels across sentences
could affect the ability to produce truly `universal'
representations that are effective for an arbitrary
classification task\footnote{To be clear, we are \textit{not}
proposing a new methodology for USEs: we are analyzing a current evaluation methodology.}.

\section{USEs for
Downstream Classification}
\label{sec:towards_usr}

Formally, USEs learn a  mapping
from a token sequence $t_1,...,t_N$ of arbitrary length $N$ to
a fixed-length vector $\vec{v} \in \mathbb{R}^n$.
When evaluated with SentEval, the $\vec{v}$s become the input representations for learning
a logistic regression classifier for a given text classification dataset.
Currently, USEs are real-valued vectors where the
dimensionality rarely exceeds 10k.
However, in light
of this evaluation methodology, we posit that a `perfect' USE would be a high-dimensional
vector where each dimension corresponds to a certain label for a certain classification task\footnote{We make the assumption
that there would be a finite set of valid classification tasks. Future work can extend
our analysis to an infinite set of tasks, though it is not immediately clear that,
theoretically speaking, this set would be infinite.}.
The vector would be all zeros unless it has a label for a class, and in which case that dimension
has a value of one. Suppose a classification task's $k$-way classification correspond
to dimensions $(d,...,d+k-1)$, then the perfect classifier would be a vector with all zeros, except for
ones in dimensions $(d,...,d+k-1)$.

We argue that a sentence is valid
if it's corresponding vector has at
least one non-zero entry. Furthermore,
stacking all sentence vectors vertically together
to form a matrix\footnote{Like with the set of task labels, we make the
assumption that there is a finite set of sentences, which is a simplification of the
problem for our analysis, since one can construct a sentence of ever-increasing
length \cite{weisler2000theory}.}, the columns now
correspond to classification tasks,
which we argue must have at least one
non-zero entry to be a valid
classification task.
Furthermore, we do not allow duplicate
rows/columns (sentences/task labels).
We define this matrix as $\mathcal{U}$.

\section{Low-Rank Approximation of USEs}
\label{sec:l-r app}

Because the goal of USEs is to be applicable to all
classification tasks, the lower-dimensional embeddings must
approximate the matrix $\mathcal{U}$.
The Eckart-Young-Mirsky theorem \cite{eckart1936approximation}
states that the solution to this problem is in fact the
singular value decomposition of $\mathcal{U}$.
Moreover, the difference
in norm between a matrix $D$ and a rank $r$ matrix $\hat{D}^*$,
derived from its singular
value decomposition, is related to the singular values of $D$,
$\sigma_{i}$, as follows:

\begin{equation}\label{eq:e-y-m_theorem}
    ||D-\hat{D}^*||_F = \sqrt{\sigma^2_{r+1}+...+\sigma^2_{m}}
\end{equation}

\noindent By construction of $\mathcal{U}$, it is full-rank and thus
has all non-zero singular values. Therefore, there is guaranteed
to be information loss when reconstructing the complete
matrix with a lower-rank approximation. Note that $\hat{D}^*$
is the optimal low-rank approximation, making the left-hand
side of Equation \ref{eq:e-y-m_theorem} a lower bound on the
reconstruction loss.

Suppose we would like to use this low-rank approximation as the
representations for a logistic regression model to perform
binary classification as to whether a sentence has a label.
The column of the right-singular
vector matrix $V$, associated with that label for
that task, will provide the best possible binary classifier for
this objective, when attempting to 
generalize across all possible input sentences
to the classifier.
Such a classifier would have the
smallest possible generalization error given USEs
of a certain dimension.

\section{Effect of Label Density}
\label{sec:eff_ld}

\begin{figure*}
\includegraphics[scale=.413]{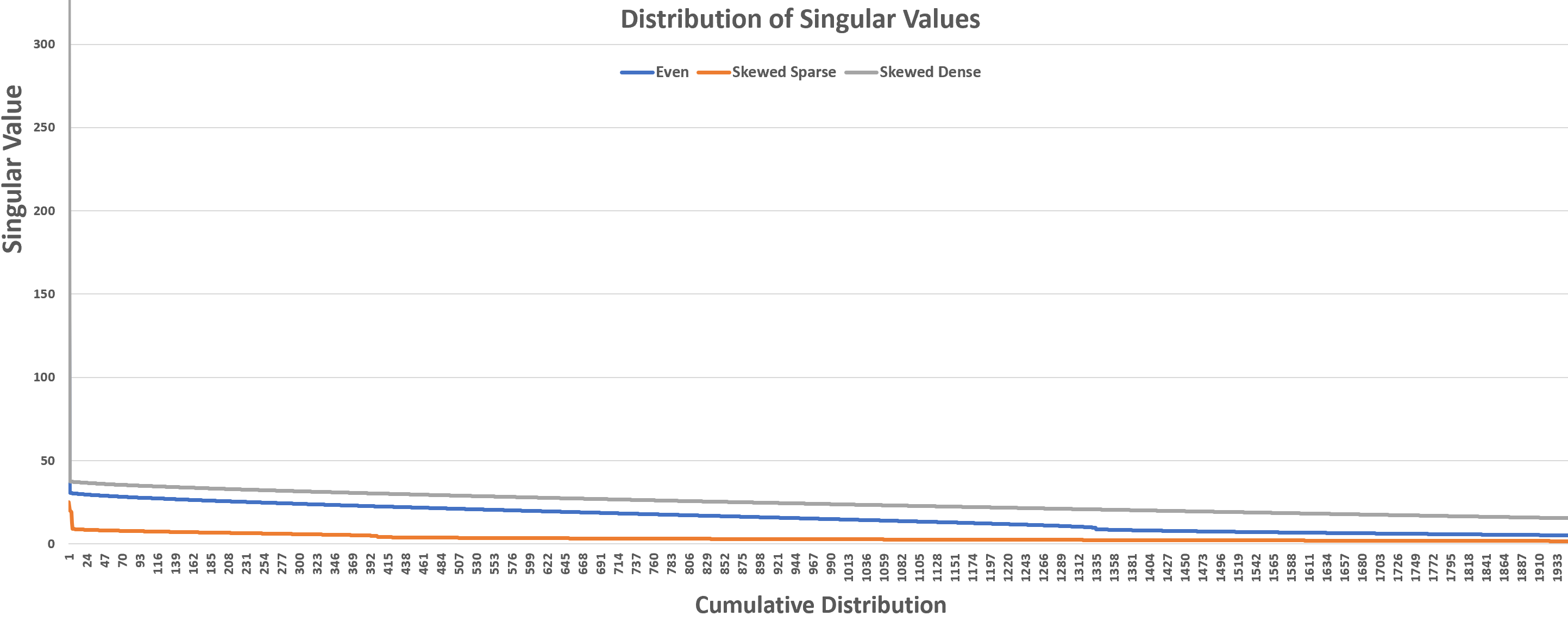}
\caption{\label{fig:sv}The cumulative distribution of the first 2k singular values for the matrices
in Section \ref{sec:eff_ld}. From
Equation \ref{eq:e-y-m_theorem}, this allows
us to visualize the information lost using a low-rank
approximation for each matrix with 40 dimensions.}
\end{figure*}

Given this connection between USEs and the low-rank
approximation of $\mathcal{U}$, it is informative
to understand the relationship between how labels
are distributed across sentences and how much error
there is using a low-rank approximation to estimate
$\mathcal{U}$. We conduct experiment on a binary
matrix, $A$ size 4,000 by 4,300,
with each experimental variation modifying the density (percentage
of non-zero elements).
Rows can either be 0.1\%, 1\%, or 10\% dense, and the
number of rows with each type of density can either
be even or skewed, with one type taking 90\% of rows,
another 9.9\%, and the last only 0.1\%.
We factor the matrix using singular value
decomposition
(the distribution of the singular values are shown in
Figure \ref{fig:sv}), and keep only the 40 first principal
dimensions. We then reconstruct $A$ using this low-rank
approximation, and calculate the l1 loss across each
row. Finally, we average the row-wise loss across each
density type. The results are shown in Table \ref{tab:row_loss}.

One of the most interesting observations is that
when the most dense rows account for the least amount
of rows in the matrix, it is actually the 
rows with a density of 1\% that lose the most information. It is also interesting to see that
the average losses comparing evenly distributed rows
to ones skewed dense are almost identical. Lastly,
and arguably most importantly,
note that the sparsest rows lose the least amount of
information. 

\begin{table}[t]
\begin{center}
\begin{tabular}{|c|c|c|}
\hline \bf Row Density & \bf \% Coverage & \bf Avg Loss \\ \hline
0.1\% & 90 & $14 \pm 5.6$\\
1\% & 9.9 & $139 \pm 34.4$\\
10\% & 0.1 & $39 \pm 1.4$\\ \hline
0.1\% & 33.3 & $15 \pm 5.2$\\
1\% & 33.3 & $92 \pm 12.4$\\
10\% & 33.3 & $750 \pm 28.4$\\ \hline
0.1\% & 0.1 & $16 \pm 5.0$\\
1\% & 9.9 & $93 \pm 12.3$\\
10\% & 90 & $754 \pm 28.5$\\
\hline



\end{tabular}
\end{center}
\caption{\label{tab:row_loss} The effect of sentence-label density on reconstruction loss using a low-rank
approximation.
See Section \ref{sec:eff_ld} for further details.}
\end{table}

\section{Experiments}
In order to further empirically determine the effect of various factorization scenarios with respect to sentence
embeddings' performance on downstream classification tasks, we ran several experiments on the SentEval
classification tasks. Each scenario has three different attributes that describe it:\\
\\
\textbf{Rep.}: We have two ways to create the representations in our experiments: 1) Start with
with a high dimensional binary matrix, and factor it with SVD. The matrix we factor is always of size
70,000 X 120,000, where the first $N$ rows map to a unique data point $d$
in a given classificaiton dataset,
and the first $k$ columns are used to determine if $d$ is of class {1,...,$k$}\footnote{We specifically \textit{include} the task label information in the representation, since we are
not evaluating a specific technique for USEs, but rather, how well USEs can encode
the information contained in $\mathcal{U}$.}.
Consequently, none of the rows
after $N$ have values in the first $k$ columns.
The representations created this way are called SVD; 2) Use binary
vectors directly. This is similar to the factored matrix from the SVD representation, except in this
scenario the matrix strictly has $N$ rows.
\\
\textbf{\# Dim}: The dimensionality of the sentence embeddings. if the representation is SVD, this is
the number of principal components that are kept. If the representation is Binary, this is the size
of the binary vector representations.
\\
\textbf{Density}: This is the probability that a given cell of the matrix has a one. For SVD, this refers
to the binary matrix that is factored. For Binary, this is for the representations themselves.\\

\begin{table*}[t!]
\begin{center}
\begin{tabular}{|c|c|c|c|c|c|c|c|c|c|}
\hline \textbf{\# Dim.} & \textbf{Rep.} & \textbf{Density} & \textbf{CR} & \textbf{MR} & \textbf{MPQA} & \textbf{SUBJ} & \textbf{SST2} & \textbf{SST5} & \textbf{TREC}\\ \hline
40,000 & Binary & 50\% & 66.01 & 75.07 & 79.57 & 73.66 & 99.95 & 45.7 & 30.6\\
4,000 & Binary & 50\% & 83.15 & 99.57 & 97.5 & 99.74 & 100 & 99.64 & 91.2\\
400 & Binary  & 50\% & 99 & 99.99 & 98.38 & 99.99 & 100 & 100 & 98.8\\
40 & Binary & 50\% & 99.92 & 100 & 97.73 & 100 & 100 & 100 & 100\\
40,000 & Binary & 1\% & 91.02 & 91.67 & 97.86 & 99.86 & 100 & 99.95 & 98.2\\
4,000 & Binary & 1\% & 88.95 & 99.82 & 97.86 & 99.86 & 100 & 99.73 & 100\\
400 & Binary & 1\% & 99.65 & 100 & 98.15 & 100 & 100 & 100 & 100\\
40 & Binary & 1\% & 99.31 & 100 & 97.73 & 100 & 100 & 100 & 99.6\\
4,000 & SVD & 50\% & 63.76 & 61.62 & 72.51 & 60.58 & 99.89 & 29.5 & 19\\
400 & SVD & 50\% & 63.76 & 54.34 & 68.77 & 52.23 & 98.35 & 24.62 & 19.2\\
40 & SVD & 50\% & 63.76 & 50.63 & 68.77 & 50.58 & 97.47 & 23.08 & 18.6\\
4,000 & SVD & 1\% & 63.76 & 99.88 & 73.5 & 99.93 & 100 & 97.24 & 98\\
400 & SVD & 1\% & 63.76 & 99.81 & 68.77 & 99.84 & 100 & 100 & 96.6\\
40 & SVD & 1\% & 63.76 & 99.98 & 68.77 & 99.8 & 100 & 51.72 & 90.4\\ \hline
\multicolumn{3}{|c|}{MTE  (4096 Dim)} & 88.3 & 82.8 & 91.3 & 94.2 & 84.5 & - & 94.2 \\ \hline
\multicolumn{3}{|c|}{\# classes} &2&2&2&2&2&5&6\\
\multicolumn{3}{|c|}{\% Coverage} &5.7&15.7&15.7&14.3&100&17.1&8.6\\
\hline
\end{tabular}
\end{center}
\caption{\label{tab:results} Results of our embedding experiments on SentEval. MTE
are the embeddings from \citet{subramanian2018learning}. \# classes refers
to how many classes a given classification dataset has. \% Coverage refers to what
percent of rows from the factored matrix (for the SVD representation) contains examples
from a given classification dataset.}
\end{table*}

\noindent Results of our experiments using various combinations
of Rep, \# Dim, and Density
are listed in Table \ref{tab:results}.
We run our experiments on the classification
tasks available in the SentEval suite
\citep{conneau2018senteval}. Specifically, we use the following
datasets: sentiment analysis (MR and both binary and 5-class SST)
\citep{pang2005seeing,socher2013recursive},
question-type (TREC) \citep{voorhees2000building},
subjectivity/objectivity (SUBJ) \citep{pang2004sentimental},
and opinion polarity (MPQA) \citep{wiebe2005annotating}.

\section{Results}
\label{sec:results}
The results of our experiments show that there
are numerous factors that effect the performance
of USEs for downstream classification tasks.
For the binary Rep type, having a large, dense
representation, even though it has indices
reserved specifically for revealing the label on
a given task, proves harder to learn a strong classifier,
with the difficulty increased by more classes
and fewer training examples. For the SVD Rep,
given either a dense or sparse binary matrix that is factored,
the performance decreases as the dimensionality
is decreased. Also for the SVD Rep, we can
seen that \% Coverage has an important effect
on performance, as the datasets with smaller
coverage of the factored matrix often have
lower performance.

For both Reps, the density
of the matrix, either binary or factored, has
a critical impact in performance across the datasets.
This dictates that data points are harder to
classify when they have more labels from
all possible classification tasks,
which aligns with the results from Section \ref{sec:eff_ld}.
This could
potentially be further supported by examining the performance
of the state-of-the-art USEs from \citet{subramanian2018learning} (MTE).
The tasks for which it performs the worst,
 CR, MR, and SST2, are very similar, common tasks,
so it can be argued the sentences in one of the
datasets are likely to have more labels overall
across all possible classification tasks, and
therefore will be harder to classify using USEs.

\section{Other Downstream Evaluation
Tasks}

Until this point, we have discussed the usage
of downstream classification tasks for evaluating
USEs. Alternatively, the task of Semantic Textual
Similarity (STS) \citep{agirre2012semeval} is another
popular downstream task for evaluating USEs
\citep{arora2016simple,cer2018universal}.
This is an unsupervised transfer task, since
the USE for a give sentence in the dataset
is used directly without training, by taking
the cosine similarity between two sentences'
USEs as a proxy STS score. Consider, now, if
the USEs were those described in Section \ref{sec:towards_usr}, which are constructed with
the goal of having optimal transferability to
an arbitrary classification task. The calculation
of the cosine similarity between two USEs $a,b$
becomes:
\begin{equation}
    cos(a,b) = \frac{\#\; shared \; labels}{TL}
\end{equation}
where $TL$ is the number of labels assigned to
sentences $a,b$ multiplied together. Thus, two sentences will
have a higher cosine similarity when they share
more labels across classification tasks. This
result provides an interesting definition for
STS: two sentences have a higher similarity score
if they share more labels from classification tasks.

A third type of downstream task is classification
on input sentence pairs, using tasks such as
paraphrase detection \cite{dolan2004unsupervised} and natural language
inference \cite{bowman2015large}. When using paired
text as input in SentEval, the engine concatenates
the pointwise multiplication of the sentences vectors
with the absolute value of their difference. Using the representations from Section \ref{sec:towards_usr}, the concatenated vectors
would be binary negations of each other, highlighting
the labels that are shared between the sentences.

\section{Conclusion}

In this work we have examined the consequences of
evaluating Universal Sentence Embeddings on
downstream classification tasks by viewing
Universal Sentence Representations as a low-rank
approximation of the sentence-task label matrix.
Our work shows that sentences with fewer labels
across all tasks have smaller reconstruction loss,
and are therefore easier to compress in a low-dimensional vector space. Furthermore,
depending on the amount of such sentences, either the most dense \textit{or} the moderately dense will have the highest loss. Future work should seek
to provide a better theoretical distribution of
labels across the sentence-task label matrix, as well as extend analysis to the infinite-dimensional
case.
\bibliography{acl2019}
\bibliographystyle{acl_natbib}

\end{document}